\documentclass[11pt,a4paper]{article}
\usepackage[hyperref]{naaclhlt2018}
\usepackage{times}
 \usepackage{latexsym}

\usepackage{hyperref}

\usepackage{etoolbox}
 \apptocmd{\thebibliography}{\raggedright}{}{}

\usepackage{amsmath, amssymb}
\usepackage{multirow}
\usepackage{microtype} 

\usepackage{graphicx}

\makeatletter
\newcommand*\bigcdot{\mathpalette\bigcdot@{.5}}
\newcommand*\bigcdot@[2]{\mathbin{\vcenter{\hbox{\scalebox{#2}{$\m@th#1\bullet$}}}}}
\makeatother

\def\nl#1{\textsl{ #1}}

\def\imagetop#1{\vtop{\null\hbox{#1}}}
\def\maxblock{{\small\textsc{MaxBlock}}}

\def\P{{\cal P}}
\def\T{{\cal T}}
\newcommand{\R}{\mathbb{R}}
\newcommand{\E}{\mathbb{E}}
\def\frac#1#2{{\begingroup #1\endgroup\over #2}}

\aclfinalcopy 
\def\aclpaperid{365} 

\title{Bootstrapping Generators from Noisy Data}

 \author{Laura Perez-Beltrachini \and Mirella Lapata \\
Institute for Language, Cognition and Computation \\
School of Informatics, University of Edinburgh \\
10 Crichton Street, Edinburgh EH8 9AB \\
   {\tt \{lperez,mlap\}@inf.ed.ac.uk}  \\}

\date{}

\begin{document}
\maketitle
\begin{abstract}
  A core step in statistical data-to-text generation concerns learning
  correspondences between structured data representations (e.g.,~facts
  in a database) and associated texts. In this paper we aim to
  bootstrap generators from large scale datasets where the data
  (e.g.,~DBPedia facts) and related texts (e.g.,~Wikipedia abstracts)
  are loosely aligned. We tackle this challenging task by introducing
  a special-purpose content selection mechanism.\footnote{Our code and data
  are available at \url{https://github.com/EdinburghNLP/wikigen}.} We use multi-instance
  learning to automatically discover correspondences between data and
  text pairs and show how these can be used to enhance the content
  signal while training an encoder-decoder architecture. Experimental
  results demonstrate that models trained with content-specific
  objectives improve upon a vanilla encoder-decoder which solely
  relies on soft attention.
\end{abstract}

\section{Introduction}

A core step in statistical data-to-text generation concerns learning
correspondences between structured data representations (e.g.,~facts
in a database) and paired texts
\cite{barzilay2005collective,jkim:coling2010,liang09semantics}.  These
correspondences describe how data representations are expressed in
natural language (\textit{content realisation}) but also indicate
which subset of the data is verbalised in the text (\textit{content
  selection}).

Although content selection is traditionally performed by domain
experts, recent advances in generation using neural networks
\cite{bahdanau2015neural,seqLevelTraining} have led to the use of
large scale datasets containing loosely related data and text pairs.
A prime example are online data sources like DBPedia \cite{auer2007dbpedia}
and Wikipedia and their associated texts which
are often independently edited. Another example are sports databases
and related textual resources.  Wiseman et
al.~\shortcite{wiseman-shieber-rush:2017:EMNLP2017} recently define a
generation task relating statistics of basketball games with
commentaries and a blog written by fans. 

\begin{figure*}[!htb]
\begin{tabular}[t]{@{}p{4.6cm}@{\hspace*{-.1cm}}p{11.6cm}}
\hspace*{-0.2ex}(a)
 \raisebox{1.5ex}[0pt]{\imagetop{\includegraphics[scale=.4]{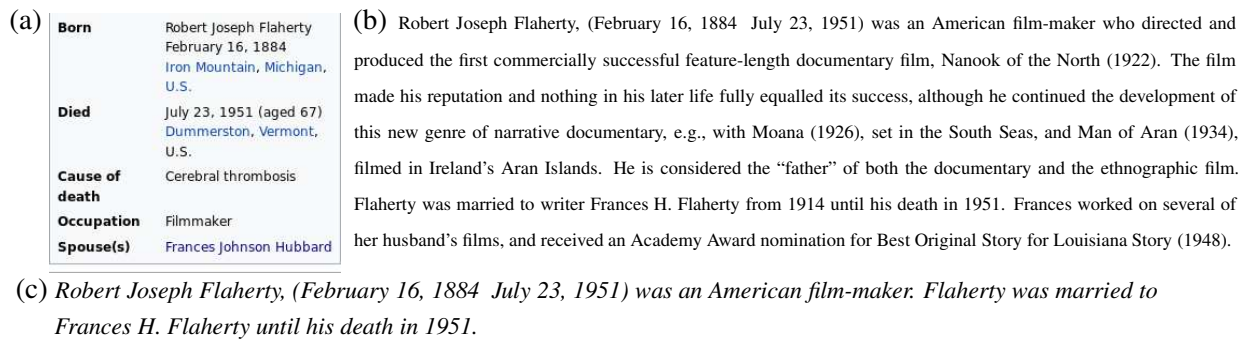}}}
&
(b)
{\scriptsize
  Robert Joseph Flaherty, (February 16, 1884 – July 23, 1951) was an American film-maker who directed and produced the first commercially successful feature-length documentary film, Nanook of the North (1922). The film made his reputation and nothing in his later life fully equalled its success, although he continued the development of this new genre of narrative documentary, e.g., with Moana (1926), set in the South Seas, and Man of Aran (1934), filmed in Ireland's Aran Islands. He is considered the ``father'' of both the documentary and the ethnographic film.
  Flaherty was married to writer Frances H. Flaherty from 1914 until his death in 1951. Frances worked on several of her husband's films, and received an Academy Award nomination for Best Original Story for Louisiana Story (1948).}
 \\
\multicolumn{2}{c}{\hspace{-1.2cm}(c)~\emph{\small Robert Joseph
    Flaherty, (February 16, 1884 – July 23, 1951) was an American
    film-maker. Flaherty was married to }}\\
\multicolumn{2}{l}{\hspace{.35cm}\emph{\small Frances H. Flaherty until his death in 1951.}}
 \end{tabular} 
\vspace*{-2ex}
\caption{Property-value pairs (a), related biographic abstract (b) for 
the Wikipedia entity \textit{Robert Flaherty}, and model verbalisation
in italics (c).
}\label{fig:looselyDataTextPair}
\end{figure*}

In this paper, we focus on short text generation from such loosely
aligned data-text resources.  We work with the biographical subset of
the DBPedia and Wikipedia resources where the data corresponds to
DBPedia facts and texts are Wikipedia abstracts about people.
Figure~\ref{fig:looselyDataTextPair} shows an example for the
film-maker \textit{Robert Flaherty}, the Wikipedia infobox, and the
corresponding abstract. We wish to bootstrap a data-to-text generator
that learns to verbalise properties about an entity from a loosely
related example text. Given the set of properties in
Figure~(\ref{fig:looselyDataTextPair}a) and the related text in 
Figure~(\ref{fig:looselyDataTextPair}b), we want to learn
verbalisations for those properties that are mentioned in the text and
produce a short description like the one in
Figure~(\ref{fig:looselyDataTextPair}c).

In common with previous work
\cite{mei2015talk,lebret-grangier-auli:2016:EMNLP2016,
  wiseman-shieber-rush:2017:EMNLP2017} our model draws on insights
from neural machine translation
\cite{bahdanau2015neural,sutskever2014sequence} using an
encoder-decoder architecture as its
backbone. \citet{lebret-grangier-auli:2016:EMNLP2016} introduce the
task of generating biographies from Wikipedia data, however they focus
on single sentence generation. We generalize the task to
multi-sentence text, and highlight the limitations of the standard
attention mechanism which is often used as a proxy for content
selection. When exposed to sub-sequences that do not correspond to any
facts in the input, the soft attention mechanism will still try to
justify the sequence and somehow distribute the attention weights over
the input representation \cite{ghader-monz:2017:I17-1}. The decoder
will still memorise high frequency sub-sequences in spite of these not
being supported by any facts in the input.

We propose to alleviate these shortcomings via a specific content
selection mechanism based on multi-instance learning (MIL;
\citeauthor{keeler1992self}, \citeyear{keeler1992self}) which
automatically discovers correspondences, namely alignments, between
data and text pairs. These alignments are then used to modify the
generation function during training. We experiment with two frameworks
that allow to incorporate alignment information, namely multi-task
learning (MTL; \citeauthor{Caruana93multitasklearning},
\citeyear{Caruana93multitasklearning}) and reinforcement learning (RL;
\citeauthor{williams1992simple}, \citeyear{williams1992simple}). In
both cases we define novel objective functions using the learnt
alignments.  Experimental results using automatic and human-based
evaluation show that models trained with content-specific objectives
improve upon vanilla encoder-decoder architectures which rely solely
on soft attention.

The remainder of this paper is organised as follows.  We discuss
related work in Section~\ref{sec:relatedWork} and describe the
MIL-based content selection approach in
Section~\ref{sec:contentSelection}. We explain how the
generator is trained in Section~\ref{sec:generation} and present
evaluation experiments in Section~\ref{sec:experimentAndResults}. 
 Section~\ref{sec:conclusion} concludes the paper.

\section{Related Work}
\label{sec:relatedWork}

Previous attempts to exploit loosely aligned data and text corpora
have mostly focused on extracting verbalisation spans for data
units. Most approaches work in two stages: initially, data units are
aligned with sentences from related corpora using some heuristics and
subsequently extra content is discarded in order to retain only 
text spans verbalising the data.  \newcite{belz2010extracting} obtain
verbalisation spans using a measure of strength of association
between data units and words,
\newcite{walter2013corpus} extract textual patterns from paths in dependency trees
while \newcite{mrabet:webnlg16} rely on crowd-sourcing. 
Perez-Beltrachini and Gardent \shortcite{perezbeltrachini-gardent:2016:*SEM} learn shared
representations for data units and sentences reduced to subject-predicate-object
triples with the aim of extracting verbalisations for
knowledge base properties.  Our work takes a step further, we not only
induce data-to-text alignments but also learn generators that produce
short texts verbalising a set of facts.

Our work is closest to recent neural network models
which learn generators from independently edited data and text resources.
Most previous work
\cite{lebret-grangier-auli:2016:EMNLP2016,chisholm-radford-hachey:2017:EACLlong,sha2017order,liu2017table}
targets the generation of single sentence biographies from Wikipedia infoboxes,
while \newcite{wiseman-shieber-rush:2017:EMNLP2017} generate game
summary documents from a database of basketball games where the input
is always the same set of table fields.  In contrast, in our
scenario, the input data varies from one entity (e.g.,~athlete) to
another (e.g.,~scientist) and properties might be present or not due
to data incompleteness. Moreover, our generator is enhanced with a
content selection mechanism based on multi-instance learning.
MIL-based techniques have been previously applied to a variety of
problems including image retrieval
\cite{maron1998multiple,zhang2002content}, object detection
\cite{carbonetto2008learning,cour2011learning}, text classification
\cite{andrews2003support}, image captioning \cite{wu2015deep,karpathy2015deep}, 
paraphrase detection \cite{xu2014extracting}, and information
extraction \cite{hoffmann2011knowledge}. The application of MIL to
content selection is novel to our knowledge.

We show how to incorporate content selection into encoder-decoder
architectures following training regimes based on multi-task learning
and reinforcement learning.  Multi-task learning aims to improve a
main task by incorporating joint learning of one or more related
auxiliary tasks. It has been applied with success to a variety of
sequence-prediction tasks focusing mostly on morphosyntax. Examples
include chunking, tagging
\cite{Collobert:ea:2011,sogaard-goldberg:2016:P16-2,bjerva-plank-bos:2016:COLING,plank:2016:COLING},
name error detection \cite{cheng-fang-ostendorf:2015:EMNLP}, and
machine translation \cite{Luong:ea:2016}. Reinforcement learning
\cite{williams1992simple} has also seen popularity as a means of
training neural networks to directly optimize a task-specific metric
\cite{seqLevelTraining} or to inject task-specific knowledge
\cite{zhang-lapata:2017:EMNLP2017}. We are not aware of any work that
compares the two training methods directly. Furthermore, our
reinforcement learning-based algorithm differs from previous text
generation approaches
\cite{seqLevelTraining,zhang-lapata:2017:EMNLP2017} in that it is
applied to documents rather than individual sentences.

\section{Bidirectional Content Selection}
\label{sec:contentSelection}

We consider loosely coupled data and text pairs where  
the data component is a set ${\P}$ of property-values 
$\{p_1:v_1, \cdots, p_{|{\P}|}:v_{|{\P}|}\}$ and the 
related text ${\T}$ is a sequence of sentences $(s_1, \cdots, s_{|\T|})$. 
We define a \emph{mention span} $\tau$ as a (possibly discontinuous)
subsequence of ${\T}$ containing one or several words that verbalise
one or more property-value from ${\P}$.  For instance, in
Figure~\ref{fig:looselyDataTextPair}, the mention span \nl{``married
  to Frances H. Flaherty''} verbalises the property-value $\{Spouse(s)
: Frances \; Johnson \; Hubbard\}$.

In traditional supervised data to text generation tasks, data units
(e.g.,~$p_i:v_i$ in our particular setting) are either covered by some
mention span $\tau_j$ or do not have any mention span at all in
${\T}$. The latter is a case of content selection where the generator
will learn which properties to ignore when generating text from such
data.  In this work, we consider text components which are
independently edited, and will unavoidably contain \emph{unaligned
  spans}, i.e.,~text segments which do not correspond to any
property-value in ${\P}$.  The phrase \nl{``from 1914''} in the text
in Figure~(\ref{fig:looselyDataTextPair}b) is such an
example. Similarly, the last sentence, talks about Frances' awards and
nominations and this information is not supported by the properties
either.  

Our model checks content in both directions; it identifies which
properties have a corresponding text span (data selection) and also
foregrounds (un)aligned text spans (text selection).  This knowledge
is then used to discourage the generator from producing text not
supported by facts in the property set~${\P}$.  We view a property
set~${\P}$ and its loosely coupled text~${\T}$ as a coarse level,
imperfect alignment. From this alignment signal, we want to discover a
set of finer grained alignments indicating which mention spans
in~${\T}$ align to which properties in~${\P}$.  For each pair $(\P,
\T)$, we learn an alignment set ${\cal A}(\P, \T)$ which contains property-value 
word pairs.  For example, for the properties $spouse$ and $died$
in Figure~\ref{fig:looselyDataTextPair}, we would like to derive the
alignments in Table~\ref{tab:exAlignments}.

\begin{table}[t]
\centering
 {\footnotesize
\begin{tabular}{ll}
 \nl{married} & $spouse: Frances Johnson Flaherty$ \\
 \nl{to} & $spouse: Frances Johnson Flaherty$\\
 \nl{Frances} & $spouse: Frances Johnson Flaherty$\\
 \nl{Flaherty} & $spouse: Frances Johnson Flaherty$\\
 \nl{death} & $died : july 23, 1951$\\
 \nl{in} & $died : july 23, 1951$\\
 \nl{1951} & $died : july 23, 1951$\\
\end{tabular}
}
\vspace*{-1.5ex}
\caption{Example of word-property alignments  for the Wikipedia
  abstract and facts in Figure~\ref{fig:looselyDataTextPair}.}\label{tab:exAlignments}
\end{table} 

We formulate the task of discovering finer-grained word alignments as
a multi-instance learning problem \cite{keeler1992self}. We assume
that words from the text are positive labels for some property-values
but we do not know which ones.  For each data-text pair~$({\P},
{\T})$, we derive $|\T|$ pairs of the form $({\P},s)$ where $|\T|$ is
the number of sentences in $\T$.  We encode property sets ${\P}$ and
sentences $s$ into a common multi-modal $h$-dimensional embedding
space.  While doing this, we discover finer grained alignments between
words and property-values. The intuition is that by learning a
high similarity score for a property set ${\P}$ and sentence pair $s$,
we will also learn the contribution of individual elements
(i.e.,~words and property-values) to the overall similarity score.  We will then use this
individual contribution as a measure of word and property-value
alignment. More concretely, we assume the pair is aligned (or
unaligned) if this individual score is above (or below) a given
threshold. Across examples like the one shown in
Figure~(\ref{fig:looselyDataTextPair}a-b), we expect the model to
learn an alignment between the text span \nl{``married to Frances
  H. Flaherty''} and the property-value $\{spouse : Frances \; Johnson
\; Hubbard\}$.

In what follows we describe how we encode $({\P}, s)$ pairs 
and define the similarity function.

\paragraph{Property Set Encoder}

As there is no fixed order among the property-value pairs $p:v$ in
${\P}$, we individually encode each one of them. Furthermore, both
properties $p$ and values $v$ may consist of short phrases. For
instance, the property $cause \; of \; death$ and value $cerebral \;
thrombosis$ in Figure~\ref{fig:looselyDataTextPair}. We therefore
consider property-value pairs as concatenated sequences $p\,v$ and use
a bidirectional Long Short-Term Memory Network (LSTM; Hochreiter and
Schmidhuber, 1997) network for their encoding. Note that the same
network is used for all pairs.  Each property-value pair is encoded
into a vector representation:
\vspace*{-1.0ex}
\begin{equation}
 \mathbf{p}_i = \text{biLSTM}_{\textit{denc}}(p\,v_i)
\end{equation}
\noindent
which is the output of the recurrent network at the final time step. We use
addition to combine the forward and backward outputs
and generate encoding $\{ \mathbf{p}_1, \cdots, \mathbf{p}_{|\P|} \}$ for~$\P$.

\paragraph{Sentence Encoder} We also use a biLSTM to obtain a
representation for the sentence $s = w_1, \cdots, w_{|s|} $.  Each
word $w_t$ is represented by the output of the forward and backward
networks at time step $t$.  A word at position $t$ is represented by
the concatenation of the forward and backward outputs of the networks
at time step $t$ :
\vspace*{-1.0ex}
\begin{equation}
 \mathbf{w}_t = \text{biLSTM}_{\textit{senc}}(w_t)
\end{equation}

\vspace*{-0.5ex}

\noindent
and each sentence is encoded as a sequence of vectors $(\mathbf{w}_1, \cdots, \mathbf{w}_{|s|})$.

\paragraph{Alignment Objective} Our learning objective 
seeks to maximise the similarity score between
property set~${\P}$ and a sentence~$s$ \cite{karpathy2015deep}. This similarity score is in
turn defined on top of the similarity scores among property-values
in~${\P}$ and words in~$s$.  Equation~(\ref{eq:PropsSentScore})
defines this similarity function using the dot product. 
The function seeks to align each word to the best scoring property-value:
\vspace{-1.0ex}
\begin{equation}\label{eq:PropsSentScore}
S_{\P s} = \sum_{t=1}^{|s|} max_{i \in \{1,\dots,|{\P}|\}} \, \mathbf{p}_i \, \bigcdot \, \mathbf{w}_t  
\end{equation}

\vspace*{-0.5ex}

\noindent
Equation~(\ref{eq:MaxMargin}) defines our objective which encourages
related properties $\P$ and sentences~$s$ to have higher similarity
than other $\P'\neq \P$ and $s' \neq s$:
\vspace{-1ex}
\begin{equation}\label{eq:MaxMargin}
\begin{split} 
{\cal L}_{CA} = max(0, S_{\P s} - S_{\P s\,'} + 1 ) \;\;\;\; \\
   +  max(0, S_{\P s} - S_{\P\,'s} + 1 ) 
\end{split}
\end{equation}

\section{Generator Training}
\label{sec:generation}

In this section we describe the base generation architecture and 
explain two alternative ways of using the alignments to guide the
training of the model.  One approach follows multi-task training where
the generator learns to output a sequence of words but also to predict
alignment labels for each word.  The second approach relies on
reinforcement learning for adjusting the probability distribution of
word sequences learnt by a standard word prediction training
algorithm.

\subsection{Encoder-Decoder Base Generator}
\label{sec:encdec}

We follow a standard attention based encoder-decoder architecture for
our generator \cite{bahdanau2015neural,luong2015effective}.  Given a
set of properties~$X$ as input, the model learns to predict an output
word sequence~$Y$ which is a verbalisation of (part of) the
input. More precisely, the generation of sequence~$Y$ is conditioned
on input~$X$:
\vspace*{-1.5ex}
\begin{equation}
 P(Y|X) = \prod_{t=1}^{|Y|} P(y_t|y_{1:t-1}, X)
\end{equation}

\vspace*{-0.5ex}
The encoder module constitutes an intermediate representation of the
input. For this, we use the property-set encoder described in
Section~\ref{sec:contentSelection} which outputs vector
representations $\{ \mathbf{p}_1, \cdots, \mathbf{p}_{|X|} \}$ for a
set of property-value pairs.  The decoder uses an LSTM 
and a soft attention mechanism \cite{luong2015effective} to
generate one word~$y_t$ at a time conditioned on the previous output
words and a context vector $c_t$ dynamically created:
\vspace*{-1.0ex}
\begin{equation}\label{eq:wordProb}
 P(y_{t+1}|y_{1:t},X) = softmax(g(\mathbf{h}_t, c_t)) 
\end{equation}

\vspace*{-0.5ex}

\noindent
where $g(\cdot)$ is a neural network with one hidden layer
parametrised by $\mathbf{W}_o \in \R^{|V| \times d}$, $|V|$ is the output vocabulary size and $d$ 
the hidden unit dimension, over $\mathbf{h}_t$ and $c_t$
composed as follows:
\vspace*{-1.0ex}
\begin{equation}\label{eq:outLayer}
 g(\mathbf{h}_t, c_t) = \mathbf{W}_o \; tanh(\mathbf{W}_c [  c_t ; \mathbf{h}_t ] )
\end{equation}
\noindent
where $\mathbf{W}_c \in \R^{d \times 2d}$. $\mathbf{h}_t$ is the hidden state of the
LSTM decoder which summarises $y_{1:t}$:
\vspace*{-1ex}
\begin{equation}
 \mathbf{h}_t = \text{LSTM}(y_t, \mathbf{h}_{t-1})
\end{equation}

\vspace*{-0.5ex}

\noindent
The dynamic context vector $c_t$ is the weighted sum of the hidden 
states of the input property set (Equation~(\ref{eq:contentVector}));
and the weights $\alpha_{ti}$ are determined by a dot product attention
mechanism:
\vspace*{-1.5ex}
\begin{equation}\label{eq:contentVector}
 c_t = \sum_{i=1}^{|X|}\alpha_{ti} \, \mathbf{p}_i
\end{equation}
\vspace*{-1ex}
\begin{equation}
\alpha_{ti} =  \frac{\text{exp}(\mathbf{h}_{t} \, \bigcdot \, \mathbf{p}_{i})}{\sum _{i^{\,\prime}} \text{exp}(\mathbf{h}_{t} \, \bigcdot \, \mathbf{p}_{i^{\,\prime}})}
\end{equation}


We initialise the decoder with the averaged sum of the encoded input
representations \cite{vinyals2015order}.  The model is trained to
optimize negative log likelihood:
\vspace*{-1ex}
\begin{equation}
 {\cal L}_{wNLL} =  - \sum_{t=1}^{|Y|} log \, P(y_t|y_{1:t-1}, X)
\end{equation}

We extend this architecture to multi-sentence texts in a way similar
to \newcite{wiseman-shieber-rush:2017:EMNLP2017}.  We view the
abstract as a single sequence, i.e.,~all sentences are concatenated.
When training, we cut the abstracts in blocks of equal size and
perform forward backward iterations for each block (this includes the
back-propagation through the encoder). From one block iteration to the
next, we initialise the decoder with the last state of the previous
block.  The block size is a hyperparameter tuned experimentally on the
development set.

\subsection{Predicting Alignment Labels}

The generation of the output sequence is conditioned on the previous
words and the input. However, when certain sequences are very common,
the language modelling conditional probability will prevail over the
input conditioning.  For instance, the phrase \nl{from 1914} in our
running example is very common in contexts that talk about periods of
marriage or club membership, and as a result, the language model will
output this phrase often,  even in cases where there are no supporting
facts in the input.  The intuition behind multi-task training
\cite{Caruana93multitasklearning} is that it will smooth the
probabilities of frequent sequences when trying to simultaneously
predict alignment labels.

Using the set of alignments obtained by our content selection model,
we associate each word in the training data with a binary label
$a_t$ indicating whether it aligns with some property in the input
set.  Our auxiliary task is to predict $a_t$ given the sequence of
previously predicted words and input $X$:
\vspace{-2ex}
\begin{equation}
  P(a_{t+1}|y_{1:t},X) = sigmoid(g'(\mathbf{h}_t, c_t)) 
\end{equation}
\vspace{-2ex}
\begin{equation}  
 g'(\mathbf{h}_t, c_t) = \mathbf{v}_a  \, \bigcdot \, tanh(\mathbf{W}_c [  c_t ; \mathbf{h}_t ] )
\end{equation}
\noindent
where $\mathbf{v}_a \in \R^{d}$ and the other operands are as 
defined in Equation~(\ref{eq:outLayer}). We optimise the following
auxiliary objective function:
\vspace{-1ex}
\begin{equation}
 {\cal L}_{aln} =  - \sum_{t=1}^{|Y|} log \, P(a_t|y_{1:t-1}, X)
\end{equation}
\noindent
and the combined multi-task objective is the weighted sum of both 
word prediction and alignment prediction losses:
\vspace{-1ex}
\begin{equation}
 {\cal L}_{MTL} = \lambda \, {\cal L}_{wNLL} + (1 - \lambda) \, {\cal L}_{aln}
\end{equation}
where $\lambda$ controls how much model training will focus on each
task.  As we will explain in Section~\ref{sec:experimentAndResults},
we can anneal this value during training in favour of one objective or
the other.

\subsection{Reinforcement Learning Training}

Although the multi-task approach aims to smooth the target
distribution, the training process is still driven by the imperfect
target text.  In other words, at each time step $t$ the algorithm
feeds the previous word $w_{t-1}$ of the target text and evaluates the
prediction against the target $w_t$.

Alternatively, we propose a training approach based on reinforcement
learning (\citealt{williams1992simple}) which allows us to define
an objective function that does not fully rely on the target text but
rather on a revised version of it. In our case, the set of alignments
obtained by our content selection model provides a revision for the
target text.  The advantages of reinforcement learning are twofold:
(a)~it allows to exploit additional task-specific knowledge
\cite{zhang-lapata:2017:EMNLP2017} during training, and (b)~enables
the exploration of other word sequences through sampling.  Our setting
differs from previous applications of RL
\cite{seqLevelTraining,zhang-lapata:2017:EMNLP2017} in that the reward
function is not computed on the target text but rather on its
alignments with the input.

The encoder-decoder model is viewed as an agent whose action space is
defined by the set of words in the target vocabulary.  At each time
step, the encoder-decoder takes action $\hat{y}_t$ with policy
$P_{\pi}(\hat{y}_t|\hat{y}_{1:t-1}, X)$ defined by the probability in
Equation~(\ref{eq:wordProb}).  The agent terminates when it emits the
End Of Sequence (EOS) token, at which point the sequence of all
actions taken yields the output sequence $\hat{Y}=(\hat{y}_1, \cdots,
\hat{y}_{|\hat{Y}|})$.  This sequence in our task is a short text
describing the properties of a given entity.  After producing the
sequence of actions~$\hat{Y}$, the agent receives a
reward~$r(\hat{Y})$ and the policy is updated according to this
reward.

\paragraph{Reward Function}
We define the reward function~$r(\hat{Y})$ on the alignment set~${\cal
  A}(X,Y)$.  If the output action sequence $\hat{Y}$ is precise with
respect to the set of alignments ${\cal A}(X,Y)$, the agent will
receive a high reward.  Concretely, we define $r(\hat{Y})$ as follows:
\vspace{-1ex}
\begin{equation}\label{eq:rewardfc}
 r(\hat{Y}) = \gamma^{pr} \, r^{pr}(\hat{Y}) 
\end{equation}
\noindent
where $\gamma^{pr}$ adjusts the reward value~$r^{pr}$ which is the
unigram precision of the predicted sequence $\hat{Y}$ and the set of
words in ${\cal A}(X,Y)$.

\paragraph{Training Algorithm}
We use the REINFORCE algorithm \cite{williams1992simple} to learn an
agent that maximises the reward function. As this is a gradient
descent method, the training loss of a sequence is defined as the
negative expected reward:
\vspace*{-1ex}
\begin{equation}
{\cal L}_{RL} = -\E_{(\hat{y}_1, \cdots, \hat{y}_{|\hat{Y}|})} \sim
P_\pi(\text{\textperiodcentered}|X)[r(\hat{y}_1, \cdots,
\hat{y}_{|\hat{Y}|})] \nonumber
\end{equation}
\noindent
where $P_\pi$ is the agent's policy, i.e.,~the word distribution
produced by the encoder-decoder model (Equation~(\ref{eq:wordProb}))
and $r(\text{\textperiodcentered})$ is the reward function as defined
in Equation~(\ref{eq:rewardfc}).  The gradient of ${\cal L}_{RL}$
is given by:
\vspace*{-1ex}
\begin{equation}
 \nabla{\cal L}_{RL} \approx \sum^{|\hat{Y}|}_{t=1}\nabla \,
 \text{log} \, P_{\pi}(\hat{y}_t|\hat{y}_{1:t-1},
 X)[r(\hat{y}_{1:|\hat{Y}|})-b_t] \nonumber
\end{equation}
\noindent
where $b_t$ is a baseline linear regression model used to reduce the
variance of the gradients during training. $b_t$ predicts the future
reward and is trained by minimizing mean squared error. The input to
this predictor is the agent hidden state $\mathbf{h}_t$, however we do
not back-propagate the error to~$\mathbf{h}_t$.  We refer the
interested reader to \citet{williams1992simple} and
\citet{seqLevelTraining} for more details.

\paragraph{Document Level Curriculum Learning}
Rather than starting from a state given by a random policy, we
initialise the agent with a policy learnt by pre-training with the
negative log-likelihood objective
\cite{seqLevelTraining,zhang-lapata:2017:EMNLP2017}.  The
reinforcement learning objective is applied gradually in combination
with the log-likelihood objective on each target block
subsequence. Recall from Section~\ref{sec:encdec} that our document is
segmented into blocks of equal size during training which we denote as
\maxblock. When training begins, only the last $\mho$ tokens are
predicted by the agent while for the first $(\text{\maxblock}-\mho)$
we still use the negative log-likelihood objective. The number of
tokens~$\mho$ predicted by the agent is incremented by $\mho$ units
every 2 epochs. We set $\mho=3$ and the training ends
when~$(\text{\maxblock}-\mho)=0$. Since we evaluate the model's
predictions at the block level, the reward function
is also evaluated at the block level.

\section{Experimental Setup}
\label{sec:experimentAndResults}

\paragraph{Data}
We evaluated our model on a dataset collated from \textsc{WikiBio}
\cite{lebret-grangier-auli:2016:EMNLP2016}, a corpus of 728,321
biography articles (their first paragraph) and their infoboxes sampled
from the English Wikipedia. We adapted the original dataset in three
ways. Firstly, we make use of the entire abstract rather than first
sentence. Secondly, we reduced the dataset to examples with a rich set
of properties and multi-sentential text. We eliminated examples with
less than six property-value pairs and abstracts consisting of one
sentence. We also placed a minimum restriction of 23 words in the length of the
abstract. We considered abstracts up to a maximum of 12~sentences
and property sets with a maximum of 50~property-value pairs. 
Finally, we associated each abstract
with the set of DBPedia properties~\mbox{$p:v$} corresponding to the
abstract's main entity. As entity classification is available in
DBPedia for most entities, we concatenate class information~$c$
(whenever available) with the property value, i.e.,~\mbox{$p:v\,c$}. 
In Figure~\ref{fig:looselyDataTextPair}, the property value $spouse :
Frances \, H. \, Flaherty $ is extended with class information from
the DBPedia ontology to $spouse : Frances \, H. \, Flaherty \, Person$.

\paragraph{Pre-processing}
Numeric date formats were converted to a surface form with month names. 
Numerical expressions were delexicalised using different tokens
created with the property name and position of the delexicalised token
on the value sequence. 
For instance, given the property-value for \nl{birth date} in 
Figure~(\ref{fig:looselyDataTextPair}a), the first sentence 
in the abstract (Figure~(\ref{fig:looselyDataTextPair}b)) 
becomes ``\nl{Robert Joseph Flaherty, (February DLX\_birth\_date\_2,
DLX\_birth\_date\_4 -- July \dots} ''.  
Years and numbers in the text not found in
the values of the property set were replaced with tokens YEAR and
NUMERIC.\footnote{We exploit these tokens to further adjust the score
  of the reward function given by Equation~\eqref{eq:rewardfc}.  Each
  time the predicted output contains some of these symbols we decrease
  the reward score by $\kappa$ which we empirically set to 0.025 .}
In a second phase, when creating the input and output vocabularies,
${\cal V}^I$ and ${\cal V}^O$ respectively, we delexicalised words~$w$
which were absent from the output vocabulary but were attested in the
input vocabulary. Again, we created tokens based on the property name
and the position of the word in the value sequence.  Words not in
${\cal V}^O$ or ${\cal V}^I$ were replaced with the symbol UNK.
Vocabulary sizes were limited to $|{\cal V}^I| = 50k$ and $|{\cal
  V}^O| = 50k$ for the alignment model and $|{\cal V}^O| = 20k$ for
the generator. We discarded examples where the text contained more
than three UNKs (for the content aligner) and five UNKs (for the
generator); or more than two UNKs in the property-value (for generation). 
Finally, we added the empty relation to the property sets.

Table~\ref{tab:datasetStats} summarises the dataset statistics for the
generator.  We report the number of abstracts in the dataset (size),
the average number of sentences and tokens in the abstracts, and
the average number of properties and sentence length in tokens 
(sent.len). For the content
aligner (cf. Section~\ref{sec:contentSelection}), each sentence
constitutes a training instance, and as a result the sizes of the
train and development sets are~796,446 and~153,096, respectively.

\setlength\tabcolsep{1.5pt} 
\begin{table}[t]
\centering
{\small
\begin{tabular}{|p{1.5cm}|c|c|c|}
 \hline
 generation  &train&dev&test\\
 \hline
 size  & 165,324 & 25,399 & 23,162 \\
 sentences & 3.51$\pm$1.99 & 3.46$\pm$1.94 & 3.22$\pm$1.72 \\
 tokens & 74.13$\pm$43.72 & 72.85$\pm$42.54& 66.81$\pm$38.16 \\
 properties & 14.97$\pm$8.82 & 14.96$\pm$8.85 & 21.6$\pm$9.97 \\
 sent.len & 21.06$\pm$8.87 & 21.03$\pm$8.85 & 20.77$\pm$8.74 \\
\hline 
\end{tabular} 
}
\vspace*{-1.5ex}
\caption{Dataset statistics.}\label{tab:datasetStats}
\end{table} 
\setlength\tabcolsep{6pt} 

\paragraph{Training Configuration}
We adjusted all models' hyperparameters according to their performance
on the development set.  The encoders for both content selection and
generation models were initialised with GloVe
\cite{pennington2014glove} pre-trained vectors. The input and hidden unit dimension
was set to 200~for content selection and 100~for
generation. In all models, we used encoder biLSTMs and decoder 
LSTM (regularised with a dropout rate of 0.3 \cite{Zaremba2014RecurrentNN})  
with one layer.
Content selection and generation models (base encoder-decoder and MTL) 
were trained for~20 epochs with the ADAM optimiser \cite{kingma2014adam} 
using a learning rate of~0.001. The reinforcement learning
model was initialised with the base encoder-decoder model
and trained for 35 additional epochs with stochastic gradient descent 
and a fixed learning rate of~0.001. Block sizes
were set to 40 (base), 60 (MTL) and 50 (RL). Weights for the
MTL objective were also tuned experimentally; we set
$\lambda=0.1$ for the first four epochs (training focuses on alignment
prediction) and switched to $\lambda=0.9$ for the remaining epochs.

\paragraph{Content Alignment} We optimized content alignment on the
development set against manual alignments. Specifically, two
annotators aligned~132 sentences to their infoboxes. We used the Yawat
annotation tool \cite{germann2008yawat} and followed the alignment
guidelines (and evaluation metrics) used in
\newcite{cohn2008constructing}. The inter-annotator agreement using
macro-averaged f-score was~0.72 (we treated one annotator as the
reference and the other one as hypothetical system output).

Alignment sets were extracted from the model's output
(cf. Section~\ref{sec:contentSelection}) by optimizing the threshold
$avg(sim) + a * std(sim)$ where $sim$~denotes the similarity between
the set of property values and words, and $a$~is empirically set to
0.75; $avg$ and $std$ are the mean and standard deviation of
$sim$~scores across the development set. Each word was aligned to a
property-value if their similarity exceeded a threshold of~0.22.  Our
best content alignment model (Content-Aligner) obtained an f-score
of~0.36 on the development set. 

We also compared our Content-Aligner against a baseline based on
pre-trained word embeddings (EmbeddingsBL).  For each pair~$({\P},s)$
we computed the dot product between words in~$s$ and properties
in~${\P}$ (properties were represented by the the averaged sum of
their words' vectors).  Words were aligned to property-values if their
similarity exceeded a threshold of~0.4. EmbeddingsBL obtained an
f-score of~0.057 against the manual alignments. Finally, we compared
the performance of the Content-Aligner at the level of property set
$\P$ and sentence $s$ similarity by comparing the average ranking
position of correct pairs among 14 distractors, namely rank@15. The Content-Aligner
obtained a rank of~1.31, while the EmbeddingsBL model had 
a rank of~7.99 (lower is better).

\section{Results}
\label{sec:results}

We compared the performance of an encoder-decoder model trained with
the standard negative log-likelihood method (ED), against a model
trained with multi-task learning (ED$_{\mathrm{MTL}}$) and
reinforcement learning (ED$_{\mathrm{RL}}$). We also included a
template baseline system (Templ) in our evaluation experiments. 

The template generator used hand-written rules to realise
property-value pairs.  As an approximation for content selection, we
obtained the 50 more frequent property names from the training set
and manually defined content ordering rules with the following
criteria. 
We ordered personal life properties (e.g., $birth\_date$ or $occupation$)
based on their most common order of mention 
in the Wikipedia abstracts. Profession dependent properties 
(e.g., $position$ or $genre$), were assigned an equal ordering but 
posterior to the personal properties.  
  We manually lexicalised properties into single sentence templates 
  to be concatenated to produce the final text. 
  The template for the property $position$ and example verbalisation
  for the property-value ${position : defender}$ 
  of the entity \nl{zanetti} are {\small ``$[$NAME$]$} \nl{played as} {\small $[$POSITION$]$.''}
  and ``\nl{Zanetti played as defender.}'' respectively.

\begin{table}[t]
\centering
{\footnotesize
 \begin{tabular}{|l|r|r|}
    \hline
  \multicolumn{1}{|c|}{Model} & \multicolumn{1}{c|}{Abstract} & \multicolumn{1}{c|}{RevAbs}  \\
    \hline
    \hline
  Templ &  5.47 & 6.43 \\
  ED &  13.46 & 35.89 \\
  ED$_{\mathrm{MTL}}$ & \textbf{13.57} & \textbf{37.18} \\
  ED$_{\mathrm{RL}}$ & 12.97 & 35.74 \\ 
      \hline
 \end{tabular} 
 }
\vspace*{-1ex}
\caption{BLEU-4 results using the original Wikipedia abstract (Abstract) as
  reference and   crowd-sourced revised abstracts (RevAbs) for 
  template baseline (Templ),  standard encoder-decoder model (ED),
  and our content-based models trained with multi-task  learning
  (ED$_{\mathrm{MTL}}$) and reinforcement learning
  (ED$_{\mathrm{RL}}$). 
}\label{tab:AutoEval}
\end{table}

\paragraph{Automatic Evaluation}

Table~\ref{tab:AutoEval} shows the results of automatic evaluation
using BLEU-4 \cite{papineni2002bleu} against the noisy Wikipedia
abstracts. Considering these as a gold standard is, however,  not entirely
satisfactory for two reasons. Firstly, our models generate
considerably shorter text and will be penalized for not generating
text they were not supposed to generate in the first place. Secondly,
the model might try to re-produce what is in the imperfect reference
but not supported by the input properties and as a result will be
rewarded when it should not.  To alleviate this, we crowd-sourced
using AMT a revised version of 200~randomly selected abstracts from
the test set.\footnote{Recently, a metric that automatically
addresses the imperfect target texts was proposed in 
\cite{dhingra-etal-2019-handling}.}

Crowdworkers were shown a Wikipedia infobox with the accompanying
abstract and were asked to adjust the text to the content present in
the infobox. Annotators were instructed to delete spans which did not
have supporting facts and rewrite the remaining parts into a
well-formed text. We collected three revised versions for each
abstract.  Inter-annotator agreement was~81.64 measured as the mean
pairwise BLEU-4 amongst AMT workers. 

Automatic evaluation results against the revised abstracts are 
also shown in Table~\ref{tab:AutoEval}.
As can be seen, all encoder-decoder based models have a significant
advantage over Templ when evaluating against both types of
abstracts. The model enabled with the multi-task learning content
selection mechanism brings an improvement of~1.29~BLEU-4 over a
vanilla encoder-decoder model.  Performance of the RL  
trained model is inferior and close to the ED model. We
discuss the reasons for this discrepancy shortly.

To provide a rough comparison with the results reported in
\newcite{lebret-grangier-auli:2016:EMNLP2016}, we also computed BLEU-4
on the first sentence of the text generated by our system.\footnote{We
  post-processed system output with Stanford CoreNLP
  \cite{manning-EtAl:2014:P14-5} to extract the first sentence.}
Recall that their model generates the first sentence of the abstract,
whereas we output multi-sentence text. Using the first sentence
in the Wikipedia abstract as reference, we obtained a score of 37.29\%
(ED), 38.42\% (ED$_{\mathrm{MTL}}$) and 38.1\% (ED$_{\mathrm{RL}}$)
which compare favourably with their best performing model
(34.7\%$\pm$0.36).

\begin{table}[t]
\begin{small}
\begin{tabular}{|@{~}l@{~}|@{~}ccccr|@{~}c@{~}|} \hline
System       & 1$^{st}$ & 2$^{nd}$ & 3$^{rd}$ & 4$^{th}$ &
       \multicolumn{1}{@{~}c@{~}|@{~}}{5$^{th}$} & Rank \\\hline\hline
Templ  & 12.17   & 14.33 & 10.17 & 15.50 & \textbf{47.83} & 3.72\\
ED&  12.83 & 24.17 & 24.67 & \textbf{25.17} & 13.17 & 3.02\\
ED$_{\mathrm{MTL}}$ & 14.83 & \textbf{26.17} & \textbf{26.17} & 19.17
& 13.67 & 2.90\\
ED$_{\mathrm{RL}}$ & 14.67 & \textbf{25.00} & \textbf{25.50} & 24.00 & 10.83 & 2.91\\
RevAbs & \textbf{47.00} & 14.00 & 12.67 & 16.17 & 9.17 & 2.27 \\ \hline
\end{tabular}
\end{small}
\vspace*{-1ex}
\caption{\label{tab:humansProportions} Rankings shown as proportions
  and mean ranks given to systems by human subjects.} 
\end{table}

\begin{table*}[t]
\begin{scriptsize}
 \begin{tabular}{|@{~}p{0.6cm}p{14.9cm}@{~}|}
  \hline
    property-set &  \textbf{name}= dorsey burnette, \textbf{date}= may 2012, \textbf{bot}= blevintron bot, \textbf{background}= solo singer, \textbf{birth}= december 28 , 1932, \textbf{birth place}= memphis, tennessee, \textbf{death place}= \{los angeles; canoga park, california\}, \textbf{death}= august 19 , 1979, \textbf{associated acts}= the rock and roll trio, \textbf{hometown}= memphis, tennessee, \textbf{genre}= \{rock and roll; rockabilly; country music\}, \textbf{occupation}= \{composer; singer\}, \textbf{instruments}= \{rockabilly bass; vocals; acoustic guitar\}, \textbf{record labels}= \{era records; coral records; smash records; imperial records; capitol records; dot records; reprise records\} \\
    RevAbs & \emph{Dorsey Burnette (December 28 , 1932 -- August 19 , 1979) was an american early Rockabilly singer. He was a  member of the Rock and Roll Trio.} \\
    Templ & \emph{Dorsey Burnette (DB) was born in December 28 , 1932. DB was born in Memphis, Tennessee. DB died in August 19 , 1979. DB died in August 19 , 1979. DB died in Canoga Park, California. DB died in los angeles. DB was a composer. DB was a singer. DB 's genre was Rock and Roll. The background of DB was solo singer. DB 's genre was Rockabilly. DB worked with the Rock and Roll Trio. DB 's genre was Country music. DB worked with the Rock and Roll Trio.} \\
    ED & \emph{Dorsey Burnette (December 28 , 1932 -- August 19 , 1979) was an american singer and songwriter. He was a member of the Rock band the band from YEAR to YEAR.} \\
    ED$_{\mathrm{MTL}}$ & \emph{Dorothy Burnette (December 28 , 1932 -- August 19 , 1979) was an american country music singer and songwriter. He was a member of the Rock band Roll.} \\
    ED$_{\mathrm{RL}}$ & \emph{Burnette Burnette (December 28 , 1932 -- August 19 , 1979) was an american singer and songwriter. He was born in memphis , Tennessee.} \\    
\hline \hline
    property-set &  \textbf{name}= indrani bose, \textbf{doctoral advisor}= chanchal kumar majumdar, \textbf{alma mater}= university of calcutta, \textbf{birth}= 1951-0-0, \textbf{birth place}= kolkata, \textbf{field}= theoretical physics, \textbf{work institution}= bose institute, \textbf{birth}= august 15 , 1951, \textbf{honours}= fna sc, \textbf{nationality}= india, \textbf{known for}= first recipient of stree sakthi science samman award \\
    RevAbs & \emph{Indrani Bose (born 1951) is an indian physicist at the Bose institute. Professor Bose obtained her ph.d. from University of Calcutta} \\
    Templ & \emph{Indrani Bose (IB) was born in year-0-0. IB was born in August 15 , 1951. IB was born in kolkata. IB was a india. IB studied at University of Calcutta. IB was known for First recipient of Stree Sakthi Science Samman Award.} \\
    ED & \emph{Indrani UNK (born 15 August 1951) is an indian Theoretical physicist and Theoretical physicist. She is the founder and ceo of UNK UNK.} \\
    ED$_{\mathrm{MTL}}$ & \emph{Indrani Bose (born 15 August 1951) is an indian Theoretical physicist. She is a member of the UNK Institute of Science and technology.} \\
    ED$_{\mathrm{RL}}$ & \emph{Indrani UNK (born 15 August 1951) is an indian Theoretical physicist. She is a member of the Institute of technology ( UNK ).} \\   
  \hline
  \hline
    property-set &  \textbf{name}= aaron moores, \textbf{coach}= sarah paton, \textbf{club}= trowbridge asc, \textbf{birth}= may 16 , 1994, \textbf{birth place}= trowbridge, \textbf{sport}= swimming, \textbf{paralympics}= 2012 \\
    RevAbs & \emph{Aaron Moores (born 16 May 1994) is a british ParalyMpic swiMMer coMpeting in the s14 category , Mainly in the backstroke and breaststroke  and after qualifying for the 2012 SuMMer ParalyMpics he won a Silver Medal in the 100 M backstroke.} \\
    Templ & \emph{Aaron Moores (AM) was born in May 16 , 1994. AM was born in May 16 , 1994. AM was born in Trowbridge.} \\
    ED & \emph{Donald Moores (born 16 May 1994) is a Paralympic swimmer from the United states. He has competed in the Paralympic Games.} \\
    ED$_{\mathrm{MTL}}$ & \emph{Donald Moores (born 16 May 1994) is an english swimmer. He competed at the 2012 Summer Paralympics.} \\
    ED$_{\mathrm{RL}}$ & \emph{Donald Moores (born 16 May 1994) is a Paralympic swimmer from the United states. He competed at the dlx\_updated\_3 Summer Paralympics.} \\
    
  \hline
  \hline 
    property-set &  \textbf{name}= kirill moryganov, \textbf{height}= 183.0, \textbf{birth}= february  7 , 1991, \textbf{position}= defender, \textbf{height}= 1.83, \textbf{goals}= \{0; 1\}, \textbf{clubs}= fc torpedo moscow, \textbf{pcupdate}= may 28 , 2016, \textbf{years}= \{2013; 2012; 2015; 2016; 2010; 2014; 2008; 2009\}, \textbf{team}= \{fc neftekhimik nizhnekamsk; fc znamya truda orekhovo- zuyevo; fc irtysh omsk; fc vologda; fc torpedo-zil moscow; fc tekstilshchik ivanovo; fc khimki; fc oktan perm, fc ryazan, fc amkar perm\}, \textbf{matches}= \{16; 10; 3; 4; 9; 0; 30; 7; 15\}\\
    RevAbs & \emph{Kirill Andreyevich Moryganov (; born 7 February 1991) is a russian professional football player. He plays for fc Irtysh Omsk. He is a Central defender.} \\
    Templ & \emph{Kirill Moryganov (KM) was born in February 7 , 1991. KM was born in February 7 , 1991. The years of KM was 2013. The years of KM was 2013. KM played for fc Neftekhimik Nizhnekamsk. KM played for fc Znamya Truda Orekhovo- zuyevo. KM scored 1 goals. The years of KM was 2013. KM played for fc Irtysh Omsk. The years of KM was 2013. KM played as Defender. KM played for fc Vologda. KM played for fc Torpedo-zil Moscow. KM played for fc Tekstilshchik Ivanovo. KM scored 1 goals. KM 's Club was fc Torpedo Moscow. KM played for fc Khimki. The years of KM was 2013. The years of KM was 2013. The years of KM was 2013. KM played for fc Amkar Perm. The years of KM was 2013. KM played for fc Ryazan. KM played for fc Oktan Perm.} \\
    ED & \emph{Kirill mikhailovich Moryganov (; born February 7 , 1991) is a russian professional football player. He last played for fc Torpedo armavir.} \\
    ED$_{\mathrm{MTL}}$ & \emph{Kirill Moryganov (; born 7 February 1991) is an english professional footballer who plays as a Defender. He plays for fc Neftekhimik Nizhnekamsk.} \\
    ED$_{\mathrm{RL}}$ & \emph{Kirill viktorovich Moryganov (; born February 7 , 1991) is a russian professional football player. He last played for fc Tekstilshchik Ivanovo.} \\      
  
  \hline
  
 \end{tabular} 
\end{scriptsize}
\vspace*{-2.5ex}
\caption{\label{tab:sys-output} Examples of system output.}
\end{table*} 

\paragraph{Human-Based Evaluation}
We further examined differences among systems in a human-based evaluation study.
Using AMT, we elicited 3 judgements for the same 200 infobox-abstract
pairs we used in the abstract revision study.  We compared the output
of the templates, the three neural generators and also included one of
the human edited abstracts as a gold standard (reference). For each
test case, we showed crowdworkers the Wikipedia infobox and five short
texts in random order. 
The annotators were asked to rank each of the
texts according to the following criteria: (1) Is the text faithful to
the content of the table?  and (2) Is the text overall comprehensible
and fluent? Ties were allowed only when texts were identical
strings. Table~\ref{tab:sys-output} presents examples of the texts
(and properties) crowdworkers saw.

Table~\ref{tab:humansProportions} shows, proportionally, how often
crowdworkers ranked each system, first, second, and so on. 
Unsurprisingly, the human authored gold text is considered best (and
ranked first 47\%~of the time). ED$_{\mathrm{MTL}}$ is mostly ranked
second and third best, followed closely by ED$_{\mathrm{RL}}$. The
vanilla encoder-decoder system~ED is mostly forth and Templ is fifth.
As shown in the last column of the table (Rank), the ranking of
ED$_{\mathrm{MTL}}$ is overall slightly better than
ED$_{\mathrm{RL}}$.  We further converted the ranks to ratings on a
scale of 1 to 5 (assigning ratings 5$\dots$1 to rank placements
1$\dots$5).  This allowed us to perform Analysis of Variance (ANOVA)
which revealed a reliable effect of system type. Post-hoc Tukey tests
showed that all systems were significantly worse
than RevAbs and significantly better than Templ ($\mbox{p $<$ 0.05}$).
ED$_{\mathrm{MTL}}$ is not significantly better than
ED$_{\mathrm{RL}}$ but is significantly ($\mbox{p $<$ 0.05}$)
different from ED.

\paragraph{Discussion}

The texts generated by ED$_{\mathrm{RL}}$ are shorter compared to the
other two neural systems which might affect \mbox{BLEU-4} scores and
also the ratings provided by the annotators.  As shown in
Table~\ref{tab:sys-output} (entity \nl{dorsey burnette}),
ED$_{\mathrm{RL}}$ drops information pertaining to dates or chooses to
just verbalise birth place information. In some cases, this is
preferable to hallucinating incorrect facts; however, in other cases
outputs with more information are rated more favourably.  Overall,
ED$_{\mathrm{MTL}}$ seems to be more detail oriented and faithful to
the facts included in the infobox (see \nl{dorsey burnette}, \nl{aaron
  moores}, or \nl{kirill moryganov}).  The template system manages in
some specific configurations to verbalise appropriate facts
(\textsl{indrani bose}), however, it often fails to verbalise
infrequent properties (\textsl{aaron moores}) or focuses on properties
which are very frequent in the knowledge base but are rarely found in
the abstracts (\textsl{kirill moryganov}).

\section{Conclusions}
\label{sec:conclusion}

In this paper we focused on the task of bootstrapping generators from
large-scale datasets consisting of DBPedia facts and related Wikipedia
biography abstracts. We proposed to equip standard encoder-decoder
models with an additional content selection mechanism based on
multi-instance learning and developed two training regimes, one based
on multi-task learning and the other on reinforcement
learning. Overall, we find that the proposed content selection
mechanism improves the accuracy and fluency of the generated texts. In
the future, it would be interesting to investigate 
a more sophisticated representation of the input \cite{vinyals2015order}.
It would also make sense for the model to decode
hierarchically, taking sequences of words \emph{and} sentences into
account \cite{zhang2014chinese,lebret2015phrase}.

\section*{Acknowledgments} 

We thank the NAACL reviewers for their constructive feedback. 
We also thank Xingxing Zhang, Li Dong and Stefanos Angelidis
for useful discussions about implementation details.
We gratefully acknowledge the financial support
of the European Research Council (award number 681760). 

\bibliography{wikidata2text_v3}
\bibliographystyle{acl_natbib}

\end{document}